\documentclass{article}
\usepackage{spconf,amsmath,graphicx}
\usepackage[latin1]{inputenc}
\usepackage{float}
\usepackage{color,soul}
\usepackage{url}

\usepackage{tabularx}
\usepackage{booktabs}
\usepackage{lipsum}
\usepackage{hhline}
\usepackage{multirow}
\usepackage{afterpage}
\usepackage[skip=5pt]{caption}
\usepackage[flushleft]{threeparttable}
\newcolumntype{b}{X}
\newcolumntype{s}{>{\hsize=.3\hsize}X}

\title{WEAKLY-SUPERVISED LOCALIZATION OF DIABETIC RETINOPATHY LESIONS IN RETINAL FUNDUS IMAGES}

\name{Waleed M. Gondal$^{\star \diamond}$\sthanks{Both authors contributed equally to this work.}\sthanks{The work of the paper was performed while the author was an intern at Bosch Center for Artificial Inelligence.}, Jan M. K\"ohler$^{\ast \star}$\sthanks{Corresponding author: jan.koehler@de.bosch.com}, René Grzeszick $^{\diamond}$, Gernot A. Fink $^{\diamond}$ and Michael Hirsch$^{\mathsection}$}

\address{$^{\star}$ Bosch
	Center for Artificial Intelligence, Robert Bosch GmbH, Stuttgart, Germany \\
    	 	$^{\diamond}$ Department of Computer Science, TU Dortmund University, Germany \\
    		$^{\mathsection}$Max Planck Institute for Intelligent Systems, Tübingen, Germany}
    
\begin{document}
\maketitle
\begin{abstract}
	
Convolutional neural networks (CNNs) show impressive performance for image classification and detection, extending heavily to the medical image domain. Nevertheless, medical experts are skeptical in these predictions as the nonlinear multilayer structure resulting in a classification outcome is not directly graspable. Recently, approaches have been shown which help the user to understand the discriminative regions within an image which are decisive for the CNN to conclude to a certain class. 
Although these approaches could help to build trust in the CNNs predictions, they are only slightly shown to work with medical image data which often poses a challenge as the decision for a class relies on different lesion areas scattered around the entire image. Using the DiaretDB1 dataset, we show that on retina images different lesion areas fundamental for diabetic retinopathy are detected on an image level with high accuracy, comparable or exceeding supervised methods. On lesion level, we achieve few false positives with high sensitivity, though, the network is solely trained on image-level labels which do not include information about existing lesions. Classifying between diseased and healthy images, we achieve an AUC of 0.954 on the DiaretDB1. 

\end{abstract}


\section{Introduction}
\label{sec:intro}
The World Health Organization (WHO) estimates that in 2002 the reason for almost 5 million blind people was diabetic retinopathy (DR), accounting for about 5\% of world blindness\footnote{www.who.int/blindness/causes/priority}. 
The estimated global prevalence of referable DR (RDR) among patients with diabetes is 35.4\%~\cite{yau2012global}. At the same time the prevalence of diabetes among adults has increased from 4.7\% in 1980 to 8.5\% in 2014 accounting for 422 million people with diabetes~\cite{world2016global}. RDR is considered to be the fifth most common cause of moderate to severe visual impairment~\cite{bourne2013causes}. Regular retinal screening for people with diabetes is recommended in order to be treated as early as possible before a moderate or severe DR has evolved leading to visual impairment. Lacking qualified personnel in developing countries \cite{resnikoff2012number} to assess retinal images, automated grading and detection algorithms have been developed. 

While first approaches using neural networks to detect diabetic retinopathy on retinal images without additional feature extraction showed a low classification accuracy \cite{gardner1996automatic,usher2004automated}, recent approaches based on deep neural networks \cite{colas2016deep,gulshan2016development,arunkumar2016multi} report good performance. For medical experts, these algorithms represent black box approaches as only a classification result but no information to why this conclusion is reached is provided. To overcome this obstacle and build trust in such automated healthcare monitoring systems, lesion areas can be detected and displayed as a basis to judge the rating.  

A lot of research has been conducted to detect specific lesion areas, like blood vessel transformations, exudates, microaneurysms and hemorrhages \cite{hatanaka2007cad,agurto2010multiscale,ravishankar2009automated,osareh2009computational,ctualu2015characterisation}. Winder et al.~\cite{winder2009algorithms} give an overview of literature from 1998-2008 using digital imaging techniques for DR. These approaches have used automatic image-processing techniques, partly combined with machine learning algorithms. Recently, lesion areas responsible for DR are detected using CNNs \cite{haloi2015improved,prentavsic2015detection}.

All these approaches have in common that specific lesion categories are detected which lead to DR but they cannot directly be connected to the prediction outcome of a deep learning algorithm. We present a method to localize areas of images which are responsible for a CNN to conclude the DR status. Though, not trained explicitly, it is shown that these areas map with the lesion areas.

\section{Related Work} 
Class-specific saliency detection in CNNs has recently received a lot of attention as it can be useful in numerous deep learning applications, e.g., in autonomous driving, where detecting a person in the scene is as important as determining its exact location in the scene. Methods for saliency map prediction identify regions which are visibly distinctive. Though, these regions may not necessarily map to areas that are decisive for image classification.

In contrast, weakly-supervised object localization corresponds to highlighting the class-specific discriminative regions which influence certain predictions. Even though CNNs can recognize the class of an object in the image, it is not easy for them to localize the object in the image. 

Recently proposed approaches \cite{zeiler2014visualizing,zhou2014object} visualize the internal representations learned by the inner layers of CNNs in order to understand their properties. In \cite{zeiler2014visualizing}, deconvolutional networks are used to visualize the patterns activated by each unit. \cite{zhou2014object} shows that while being trained to recognize scenes, CNNs learn object detectors. It demonstrates that the same network can perform both scene recognition and object localization in a single forward-pass. 

In \cite{simonyan2013deep}, class-specific maps are constructed by identifying the pixels that are most useful to predict the classification score and then back-projecting the corresponding information. Another approach mentioned in \cite{bazzani2016self} tries to identify the regions causing maximal activations while masking different portion of the images. 

In \cite{oquab2015object}, the last fully connected layers are treated as convolutions and a max pooling is applied to localize the object. The localization is limited to a point lying in the boundary of the object. Based on the similar approach, \cite{zhou2016learning} proposes class activation maps (CAMs) claiming to identify the complete extent of the object instead of one point. They use global average pooling (GAP) to leverage the linear relation between the softmax predictions and the final convolutional layer, which results in highlighting the most discriminative image regions relevant to the predicted result. A recent comparison of three localization methods is given in \cite{samek2016evaluating}.

Object localization on retina images poses a challenge as the lesion areas - among others small red dots, microaneurysms, hemorrhages - responsible for diabetic retinopathy are scattered around the image and are often not localized within a few image regions. To the best of our knowledge, \cite{quellec2017deep} is the only approach proposed so far to detect the lesion areas within retina images which is trained in weakly-supervised fashion using only image-level labels to conclude the lesion areas. Using a generalization of backpropagation, an ensemble of CNNs is learned in which each CNN excels in the detection of a certain lesion type.

\section{Method}
\label{sec:method}

\label{sec:majhead}
This section describes our proposed deep learning approach for localizing discriminative features in DR. 

The aim is to learn a representation that enables localization of discriminative features in a retina image while at the same time achieving good classification accuracy on the same. Our proposed CNN architecture is able to highlight decisive features in a single forward pass which facilitates medical diagnosis through visual inspection. Since good class-specific features and high classification accuracy are key, we adopted the award-wining CNN architecture \emph{o\_O solution} by Antony and Brüggemann~\cite{kaggle_oo}. 

\begin{figure}
	\centering
	\includegraphics[scale=0.26]{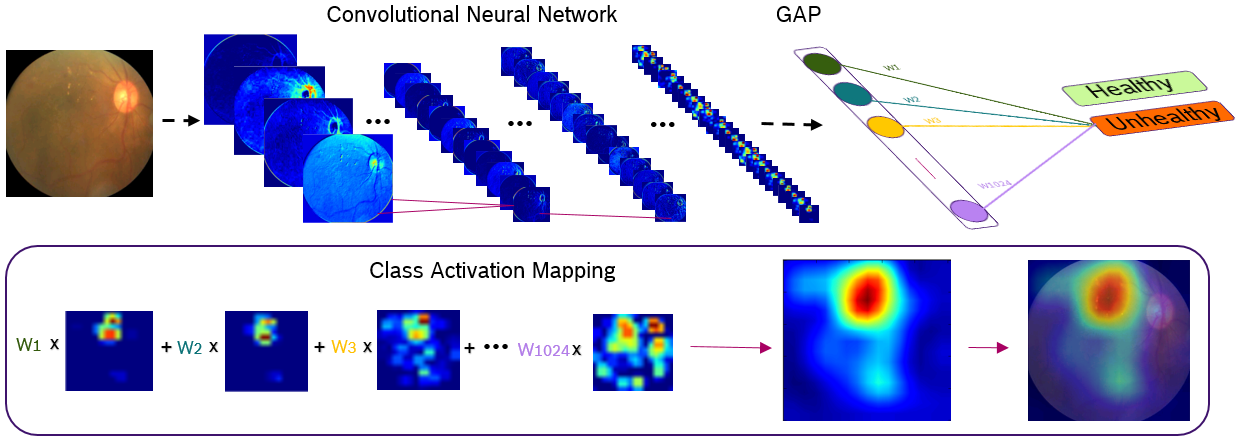}
	\vspace{-1em}
	\caption{CNN setup for generating CAMs.}
	\label{fig3}
\end{figure}

\subsection{Localization with Class Activation Maps}
The architecture has been designed to achieve good image level classification accuracy in DR. To make it capable of weakly-supervised localization, we modify it to compute CAMs introduced by \cite{zhou2016learning}. The final dense layers are removed from the proposed CNN architecture in order to retain spatial information and replaced by a GAP layer instead. The GAP layer performs average pooling on $K$ feature maps of the last convolutional layer, $A\textsuperscript{k} \in {R\textsuperscript{u x v}}$ having width $u$ and height $v$. The resultant spatially pooled values are then fully connected to output classification scores $y\textsuperscript{c}$ via $\omega_{k}^{c}$, where $c$ corresponds to the classes.

\begin{align}
	y^{c} = \sum_{k}^{} \omega_{k}^{c} \sum_{x}^{}\sum_{y}^{} 
	A_{xy}^{k}
\end{align} 
The weights $\omega_{k}$ learned in the last layer encode the importance of each feature map $A^k$ with respect to the class $c$. The final localization map $L_{CAM}$ is produced by computing the weighted linear sum of these feature maps
\begin{align}
	L_{CAM} = \sum_{k}^{} \omega_{k}^{c} A^{k}.
\end{align}
The localization map is then upsampled to the size of original input image, highlighting the class-specific image regions. The generation of class activation maps is depicted in Fig. \ref{fig3}.\\

\noindent \textbf{Fine-tuning of CAM: }Most DR lesions are of extremely small size on typical retina images. CAMs perform well in detecting those regions, but the upsampled localization map tends to produce coarse heatmaps rendering a fine-grained resolution impossible. To refine the localization map, as hinted by \cite{zhou2016learning}, the spatial resolution of the feature maps $A^k$ from the final convolutional layer is increased. In our network, we remove strides from the first and third convolutional layers, resulting in the feature maps $A^k$ of resolution $32 \times 32$ pixels. Moreover, a new convolutional layer of dimension $3 \times 3$ pixels and stride one with 1024 kernels is added to the network. 
\begin{figure}[ht!]
	\centering	
	\includegraphics[scale=0.202]{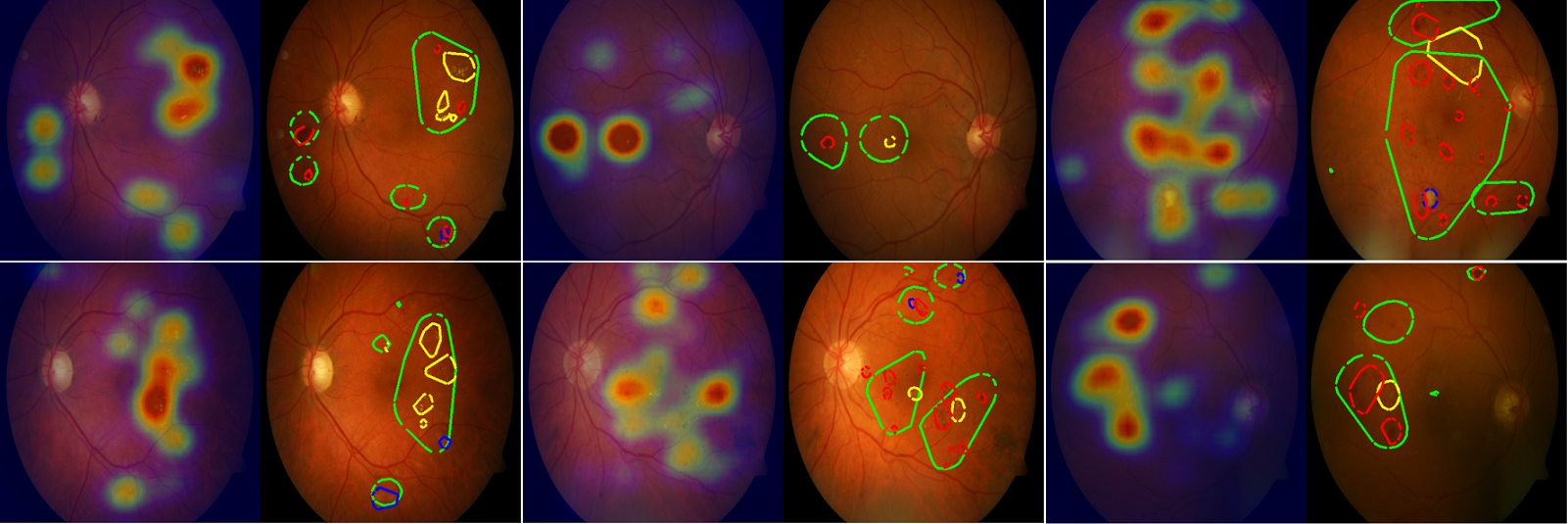} 
	\caption{The weakly-supervised localization results on DiaretDB1 images. In each pair of images, the left image shows the input image overlaid with a corresponding localization heatmap, highlighting RDR affected regions. The right image compares our detection boundary (in green) with the ground truth: yellow, blue and red marked regions represent exudates, red lesions and hemorrhages respectively. Please note that this figure is best viewed on screen rather than on print.}
	\label{fig:heatmaps}
\end{figure}
These modifications improve the overall localization ability of the network.\\

\noindent \textbf{Improving Classification Accuracy: }The removal of dense layers from the network leads to a decrease in the overall classification accuracy of the network. We also observed that increasing the spatial resolution of feature maps $A^k$ slows down the training process, significantly. At the same time, the introduction of batch normalization \cite{ioffe2015batch} in each convolutional layer during the training process enabled us to achieve faster training convergence with higher learning rates. We also employ regularization within our network to avoid over-fitting and for making our model more generic for RDR recognition and lesion localization. This is helpful since the dataset for the localization task, DiaretDB1, and the dataset used for training, were taken with different appliances~\cite{kauppi2007diaretdb1, cuadros2009eyepacs}~(see section ~\ref{sec:datasets}).

\subsection{Generation of Region Proposals}

As shown in Fig.~\ref{fig:heatmaps}, CAMs generate heatmaps highlighting class-specific discriminative regions. Heatmaps are good for qualitative analysis of the approach. However, for the evaluation of the localization results, well defined region proposals are required. To achieve this, heatmaps are first normalized between 0 and 1, assigning each pixel a value according to its intensity. The high intensity regions are then selected using binary segmentation, giving us the predicted regions $P_{i}$ for RDR lesion areas, where $i$ enumerates the predicted regions. We empirically found that a threshold value of 0.65 yields good regions. For each $P_{i}$ obtained, max-pooling is performed to get one score $S_{i}$ which serves as the confidence measure of prediction for each $P_{i}$.

\section{Experiments}
This section describes the datasets, experiments and their detailed comparison with other methods.
\subsection{Datasets}\label{sec:datasets}
Two publicly available datasets, \textit{Kaggle Diabetic Retinopathy Dataset} and \textit{DiaretDB1}, were used for this study. We use the Kaggle dataset for training and evaluate our lesion localization approach on the dataset \textit{DiaretDB1} \cite{kauppi2007diaretdb1}.\\ 

\noindent \textbf{Kaggle Dataset:} The dataset~\cite{kaggle_challenge} provided by EyePACS \cite{cuadros2009eyepacs} contains 88,702 color fundus images of which 80\% were used for training and 20\% for validation. For classification, the first two classes of the five DR levels were grouped into non-referable DR (NRDR) and the remaining three classes into RDR. In our experiments, we were more concerned with improving the network's lesion level detections performance than improving the classification accuracy, where people have already achieved remarkable results.
\\[2mm]
\noindent \textbf{DiaretDB1 Dataset:} This dataset is used to validate our lesion level detections. The dataset contains 89 color fundus images, hand-labeled by four experts for four different DR lesion types \cite{kauppi2007diaretdb1}. As suggested in \cite{kauppi2007diaretdb1}, we only consider those pixels as ground truth whose confidence level of labeling exceeds an average of 75\% between experts. 

\subsection{Implementation}
In the kaggle dataset, retina sphere is surrounded by black margins containing no information. These black regions were cropped and the images were resized to $512 \times 512$ pixels. All training images were individually standardized by subtracting mean and dividing by standard deviation which were computed over all the pixels in an image. In addition to random brightness and contrast enhancements, the images were randomly rotated, flipped horizontally and vertically in data augmentation performed during training.
The network was implemented using Tensorflow and trained on Tesla K80 GPU for 150 epochs. Gradient descent optimizer was used with the momentum of 0.8. L2 regularization was performed on weights with weight decay factor of 0.0005. The initial learning rate was 0.01 which was decayed by 1\% after each epoch.

\subsection{Evaluation on DiaretDB1 Lesion Detection} 
We assess performance at both image and lesion level.

\begin{table*}
	\caption{Lesion level performance comparison with different methods.}
        \label{lesion_level}
	\begin{tabularx}{\textwidth}{bssssssss}
		\toprule
		Method & \multicolumn{2}{c|}{\textbf{Hemorrhages}} & \multicolumn{2}{c|}{\textbf{Hard Exudates}} &
		\multicolumn{2}{c|}{\textbf{Soft Exudates}} & \multicolumn{2}{c}{\textbf{RSD}}  \\		                                         
		\hline
		& \textbf{SE\%} & \textbf{FPs/I} & \textbf{SE\%} &\textbf{FPs/I} & \textbf{SE\%} & \textbf{FPs/I} & \textbf{SE\%} &\textbf{FPs/I}  \\
		\midrule
		Quellec \textit{et al. } \cite{quellec2017deep}  & 71  & 10   & \textbf{80}  & 10  & \textbf{90} & 10 & \textbf{61} & 10 \\ 
		Dai \textit{et al. } \cite{dai2016retinal}  & -  & -   & -  & -  & - & - & 29 & 20.30 \\   
		Ours (50\% Overlap)           & \textbf{72}   & \textbf{2.25}   & 47   & \textbf{1.9} & 71 & \textbf{1.45} & 21 & \textbf{2.0} \\ \hline
		Ours (OnePixel Overlap)  & 91   & 1.5   &  87  & 1.5 & 89 & 1.5 & 52 & 1.5  \\ 
		\bottomrule
	\end{tabularx}
\end{table*}

\subsubsection{Performance at Image Level}
Most of the studies done on RDR lesion detection at image level have not mentioned their criteria for selecting true positives (TP) and false negatives. Therefore, for the sake of clarity we evaluated our approach for two scenarios. In the first scenario, an image is considered to be TP for a lesion, if there is a minimum overlap of 50\% between $P_{i}$ and the corresponding lesion's ground truth $G_{j}$, where $j$ is the number of ground truth annotations. In the second scenario an overlap of one pixel, whose confidence level is 0.75 or more, between $P_{i}$ and $G_{j}$ is considered to be a TP. Although the first criteria is more strict than the second, our method performs similar in both scenarios, ascertaining the precision of our approach. 

Our CNN based model is trained to perform binary classification on RDR, achieving 93.6\% sensitivity and 97.6\% specificity on DiaretDB1 dataset with area under the Receiver Operating Characteristics (ROC) curves of 95.4\%. For lesion level detection at image level we only report sensitivity as the pixel-wise comparison of $P_{i}$ with $G_{j}$ is possible to confirm the presence of certain RDR finding. However, if the model wrongly classifies a healthy image to be unhealthy, which is clearly a false positive (FP) at image level, it is not possible to relate this FP to any specific RDR lesion type. Thus, we only report the specificity over all lesion types which is 97.6\%.

Given that our model is trained in weakly-supervised fashion for classifying RDR, it is remarkable that it performs comparable or even outperforms fully supervised methods for image level lesion detections which are trained specifically for detecting one or two types of lesions. 
\begin{table} [t!]
	\caption{Image level sensitivity in \%. \label{image_level1}} 
	\centering 
	\begin{threeparttable}
		\resizebox{\columnwidth}{!}{%
			\begin{tabular}{c*{5}{>{}c<{}}} 
				\hline 
				Method & H\tnote{*} & HE\tnote{*} & SE\tnote{*} & RSD\tnote{*}  \\ 
				\hline 
				Zhou \textit{et al.}\cite{zhou2016automatic}& 94.4 & - & - \\ 
				Liu \textit{et al.}\cite{liu2017location} &- & 83.0 & \textbf{83.0} & - \\
				Haloi \textit{et al.}\cite{haloi2015gaussian} & & \textbf{96.5} & - & - \\
				Mane \textit{et al.}\cite{mane2015detection}  & -& - & - & \textbf{96.4} \\
				Ours (50\% Overlap) & \textbf{97.2} & 93.3 & 81.8 & 50 \\ \hline
				Ours (OnePixel Overlap) & 97.2 & 100 & 90.9 & 50  \\  
				\hline 
			\end{tabular}
		}
		\begin{tablenotes}
			\item[*] H, HE, SE, RSD: Hemorrhages, Hard Exudates, Soft-\\ Exudates and Red Small Dots.
		\end{tablenotes}	
	\end{threeparttable}
\end{table}
The comparison of sensitivities is given in Tab. \ref{image_level1}.

\subsubsection{Performance at Lesion Level}

Free-response Receiver Operating Characteristic (FROC) curves \cite{chakraborty1989maximum} are commonly used for lesions localization evaluation in medical imaging. In our evaluations only those regions $G_{j}$ which have an overlap of  at least 50\% with a $P_{i}$  are considered TP. Sometimes, $P_{i}$ are way bigger than $G_{j}$, possibly covering one or more $G_{j}$, therefore, in order to penalize this, mean Intersection over union (mIOU) for each $P_{i}$ with covered $G_{j}$  is computed. The $P_{i}$ is considered FP if its mIOU value is less than 0.5.
\begin{figure}
	\centering
	\includegraphics[scale=0.375]{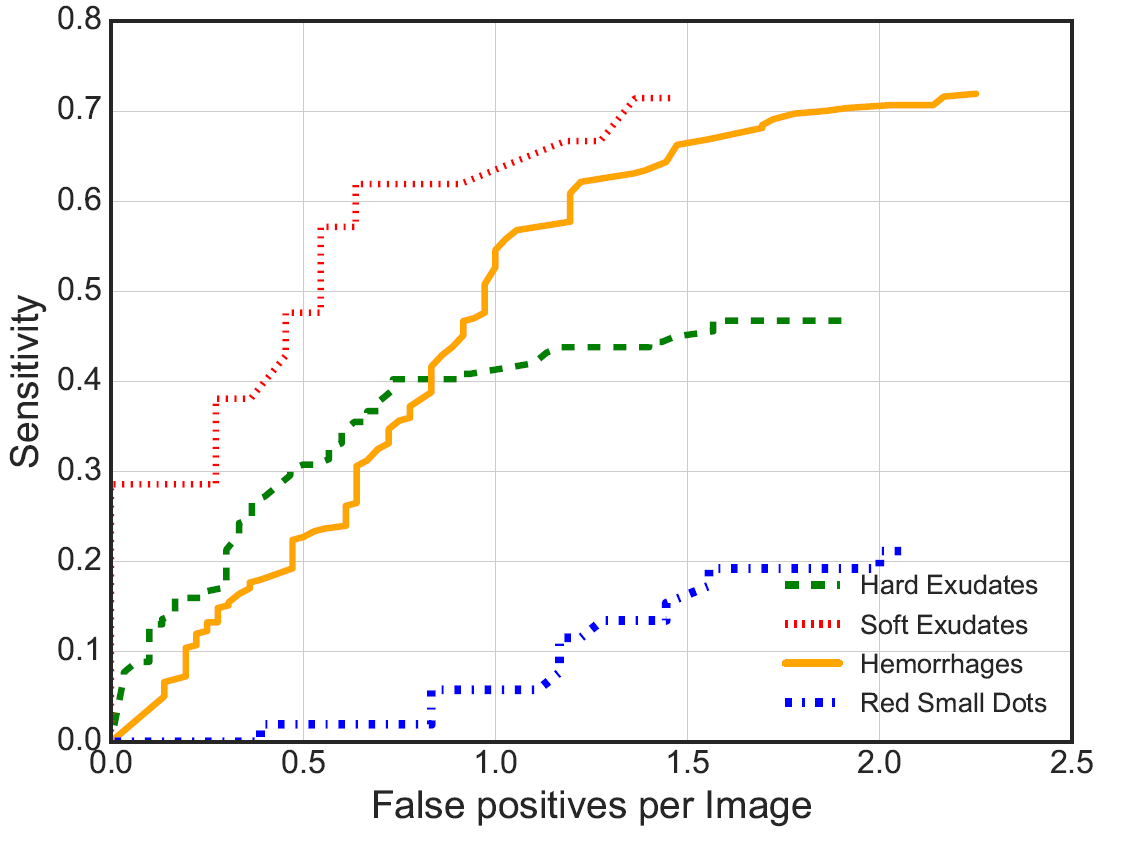}
	\caption{FROC curves for all  four types of DR lesions.}
	\label{fig2}
\end{figure}\\
Our network does not perform well in detecting red small dots which are often one or two pixels wide on a $512 \times 512$ image. We suspect that this could be due to the architecture of CNNs where the information is compressed down the stream for inference, resulting in the loss of information for very small regions. Moreover the resolution of heatmaps is too coarse to highlight these small regions precisely. We compare our localization results with the method from \cite{quellec2017deep} which employed CNN based weakly-supervised localization scheme in detecting RDR lesions. The comparison provided in Tbl.~\ref{lesion_level} shows that we have fewer FPs than other state of the art methods while achieving comparable results on sensitivity (SE). FROC plots are shown in Fig.~\ref{fig2}.

\section{Conclusion}
\label{sec:conclusion}
We presented a deep learning approach that highlights regions on retinal images that are indicative for diabetic retinopathy to assist medical diagnosis. Our architecture is inspired by a recent top-performing supervised CNN architecture for diabetic retinopathy classification but modified to enable weakly supervised object localization. We demonstrate accurate localization with good sensitivity while maintaining high classification accuracy. Along with fast inference we hope that our approach will facilitate diagnostic inspection and be a useful tool for medical professionals.

\clearpage 

{\footnotesize
\bibliography{bibliography_items}}

\bibliographystyle{My_IEEEbib}

\end{document}